%% file: aaai22.tex
\relax
\documentclass[letterpaper]{article} 
\usepackage{aaai22}  
\usepackage{times}  
\usepackage{helvet}  
\usepackage{courier}  
\usepackage[hyphens]{url}  
\usepackage{graphicx} 
\usepackage{natbib}  
\usepackage{caption} 
\DeclareCaptionStyle{ruled}{labelfont=normalfont,labelsep=colon,strut=off} 
\usepackage{algorithm}
\usepackage{algorithmic}
\usepackage[T1]{fontenc}

\usepackage{times}
\usepackage{epsfig}
\usepackage{graphicx}
\usepackage{amsmath}
\usepackage{amssymb}
\usepackage{CJKutf8}
\usepackage{adjustbox}
\usepackage{subfigure}

\usepackage{multirow}
\usepackage{hhline}
\usepackage{makecell}
\usepackage{xcolor}
\usepackage{color}

\usepackage{newfloat}
\usepackage{listings}

\floatstyle{ruled}
\newfloat{listing}{tb}{lst}{}
\floatname{listing}{Listing}

\pdfoutput=1

\title{Learning from Data with Noisy Labels Using Temporal Self-Ensemble}

\author{Jun Ho Lee,
        Jae Soon Baik,
        Tae Hwan Hwang,
        and~Jun Won Choi
}
\affiliations{Hanyang University\\
    \{jhlee, jsbaik, thhwang\}@spa.hanyang.ac.kr, junwchoi@hanyang.ac.kr
}

\usepackage{bibentry}

\begin{document}
\pdfoutput=1

\maketitle

\begin{abstract}
There are inevitably many mislabeled data in real-world datasets. Because deep neural networks (DNNs) have an enormous capacity to memorize noisy labels,  a robust training scheme is required to prevent labeling errors from degrading the generalization performance of DNNs. Current  state-of-the-art methods present a co-training scheme that trains dual networks using samples associated with small losses. In practice, however, training two networks simultaneously can burden computing resources.  In this study, we propose a simple yet effective robust training scheme that operates by training only a single network. During training, the proposed method generates {\it temporal self-ensemble} by sampling intermediate network parameters from the weight trajectory formed by stochastic gradient descent optimization. The loss sum evaluated with these self-ensembles is used to identify incorrectly labeled samples. In parallel, our method generates multi-view predictions by transforming an input data into various forms and considers their agreement  to identify  incorrectly labeled samples. By combining the aforementioned metrics, we present the proposed {\it self-ensemble-based robust training} (SRT) method, which can  filter  the  samples  with  noisy  labels to reduce their influence on training. Experiments on widely-used public datasets demonstrate that the proposed method achieves a state-of-the-art performance in some categories without training the dual networks.
\end{abstract}

\section{Introduction}

Data annotation is a labor-intensive and expensive process. Because humans can make mistakes, stringent disciplinary actions such as double-confirmation and cross-checking are required to reduce annotation errors. Modern deep neural networks (DNNs) require large training datasets to determine hundreds of thousands of model parameters.  Because accurate annotations on large datasets need a large amount of efforts and expense, various cost-effective methods such as crowdsourcing and web annotation have been adopted as alternatives. As these methods inevitably introduce labeling errors, it is essential to develop a robust training strategy to minimize the impact of mislabeled examples on test performance. Unfortunately, standard DNN training methods  have not been successful in handling such labeling errors because DNNs have a large capacity to memorize noisy labels \cite{arpit2017closer,zhang2016understanding}. Thus, numerous studies have attempted to reduce the influence of noisy labels on DNN training.  

\begin{figure*}[t] 
    \centering
    \includegraphics[width=0.95\textwidth]{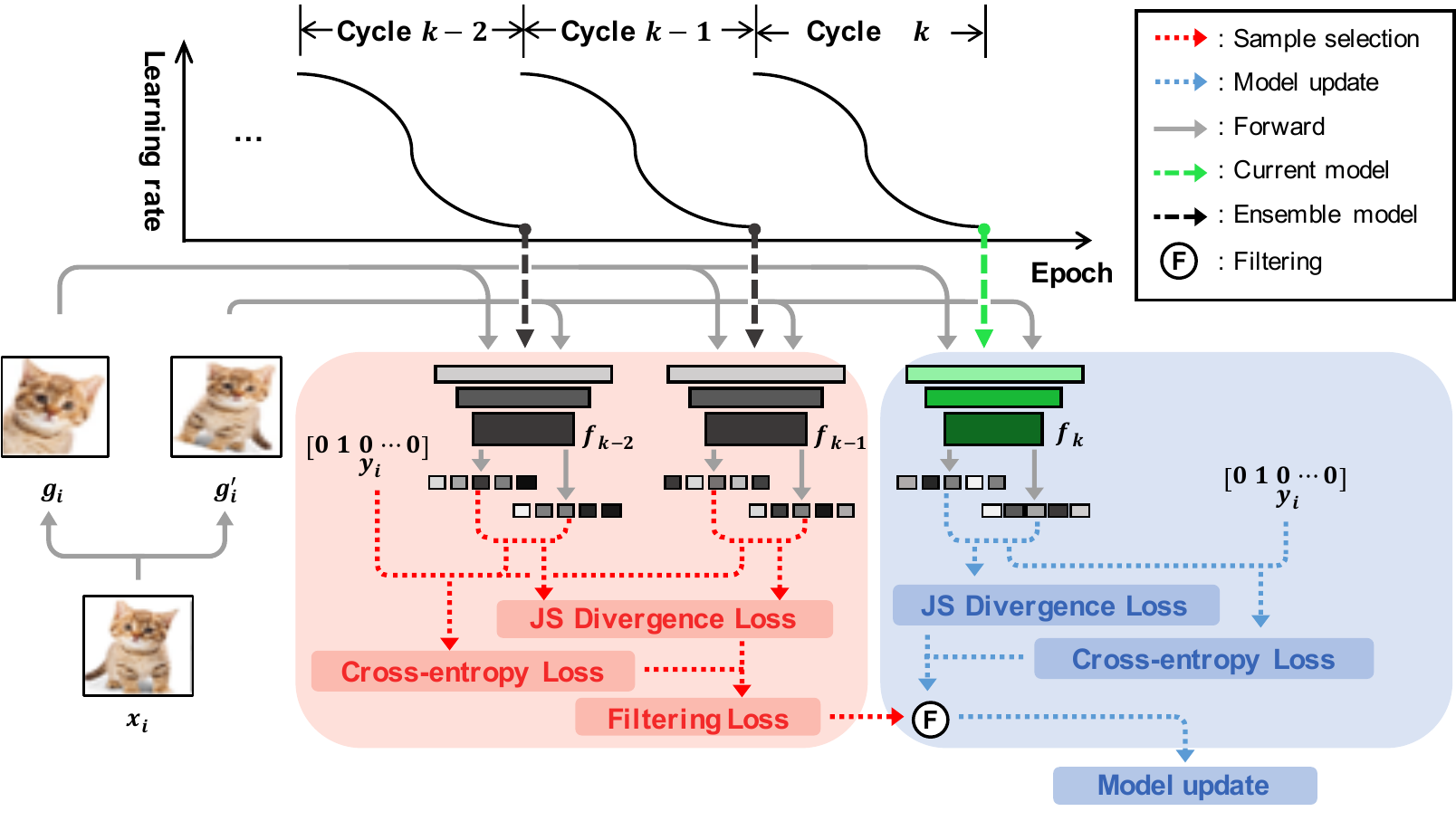}
    \caption{{\bf Overall structure of the proposed SRT method:} Temporal self-ensemble models generated using the cyclical learning rate are used to filter the noisy labels. The filtering criterion is based on the cross-entropy loss and the Jensen-Shannon divergence between multi-view predictions  evaluated with the previously generated self-ensemble models. The training is performed on the filtered samples using the loss function derived from the current model.}
    \label{fig:overall_framwork}
\end{figure*}

One approach to robust training against noisy labels is to regularize the model by weighting the loss terms for samples that possess noisy labels \cite{patrini2017fcorrection,ren2018learning,liu2015classification,wang2017multiclass,reed2014bootstrap}. Because this approach uses  noisy labels throughout the training phase, it is difficult to prevent memorization of noisy labels by DNNs. Therefore, many recent studies have proposed  strategies for filtering noisy labels during training \cite{wei2020jocor,jiang2018mentornet,malach2017decoupling,bo2018coteaching,yu2019coteaching+,song2019selfie,lee2020ltec,shen2019itlm,kim2019nlnl,yao2021josrc}. The objective of this approach is to identify the samples whose labels appear to be correct and use them for training. One promising solution is to consider the samples associated with small losses as clean samples. However, if we use the loss function used to train the network for filtering, some noisy labels memorized by the DNNs may not be filtered in the subsequent training phase. We call this {\it self-bias issue} or {\it self-confirmation issue} because model training and filtering of noise labels influence each other to degrade performance.  To overcome this self-bias issue, Co-teaching \cite{bo2018coteaching} trains two independent networks simultaneously and updates the weights of one network with the small loss samples found by the other network. Initialized with the different weights, two  networks can be co-trained in a decoupled manner, thereby reducing the effect of memorization. Decoupling \cite{malach2017decoupling} and Co-teaching+ \cite{yu2019coteaching+} further strengthen such a decoupling effect by updating their weights with the training samples whose predictions from two networks do not match.

The benefit of deploying multiple networks for a robust training is also justified by  {\it ensemble learning}. A model ensemble captures the distribution of model parameters given a training dataset. Hence, the distribution of  predictions from the model ensemble  can indicate whether a particular sample has a noisy label. This can be analogous to a case in which we can make a better decision if we listen to the opinions of different experts. Because the experts' opinions are more likely to agree on clean labels than on noisy labels, we can judge whether labels are noisy depending on their consensus. SELF \cite{nguyen2020self} formed a model ensemble by taking a moving average of the weight checkpoints obtained over  stochastic gradient descent (SGD) iterations.  LEC \cite{lee2020ltec} produced the ensemble networks by injecting perturbations into the model parameters.  JoCoR \cite{wei2020jocor} incorporated consensus measures among jointly trained networks into the loss function and used small loss samples for training. O2U-Net \cite{huang2019o2u} performed filtering based on the loss averaged over multiple cycles of cyclical learning rate.  Recently, Jo-SRC \cite{yao2021josrc} proposed the strategy of relabeling noisy samples using an ensemble model generated by an exponential moving average of the network weights.

In this paper, we propose an effective self-training scheme, that filters incorrectly labeled samples during training. The proposed method, referred to as {\it self-ensemble-based robust training} (SRT), generates temporal self-ensemble networks while training a single network with a stochastic gradient optimization. We adopt the learning rate scheduling used to generate self-ensemble models for {\it stochastic weight averaging} (SWA) \cite{izmailov2018averaging}, which was used to find an improved inference model through ensemble averaging. We build a sample acquisition function based on these self-ensembles to evaluate the likelihood of being incorrectly labeled for a given sample. The proposed acquisition function comprises two terms. First, the cross-entropy loss term is obtained using the temporal self-ensemble networks previously generated for the latest $P$ learning rate (LR) cycles. The second contrastive loss term indicates the consensus between the {\it multi-view predictions} produced by feeding the differently transformed input samples into the temporal self-ensemble networks. In addition, the multi-view predictions obtained for the  model under training are  used to enforce the consistency of  model predictions through the contrastive loss.
Fig. \ref{fig:overall_framwork} illustrates the proposed SRT method.

 We conduct the experiments on several widely used public datasets. Our results demonstrate that the filtering criterion derived from both temporal self-ensemble and multi-view predictions can offer significant performance gains over the baselines and the proposed method can achieve state-of-the-art performance for some categories.

\begin{figure*}[t]
    \centering
    \includegraphics[width=0.95\textwidth]{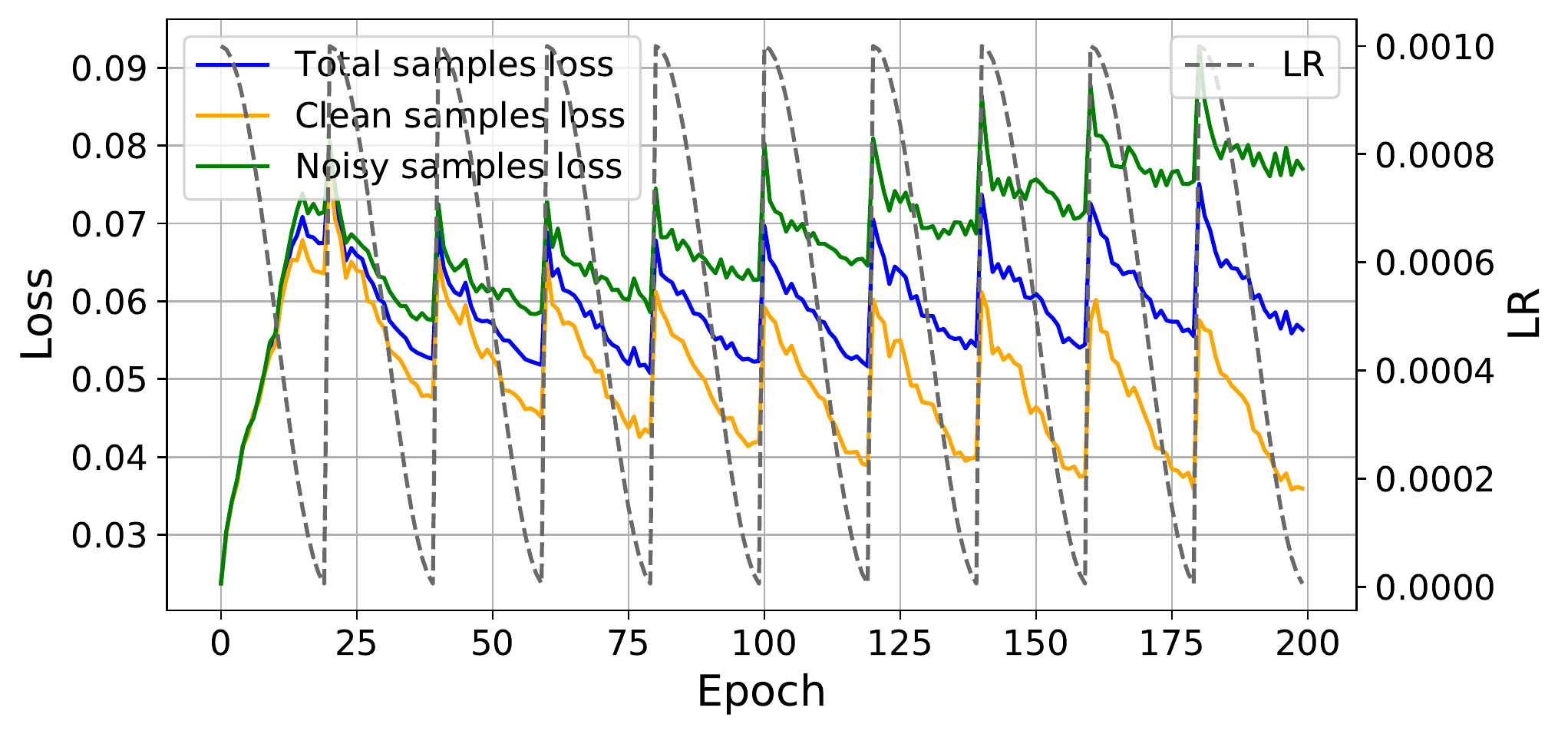}
    \caption{JS loss vs. epoch averaged over clean samples and noisy samples for each epoch evaluated on CIFAR-10 dataset.}
    \label{fig:js}
\end{figure*}


The contributions of our work are summarized as follows.
\begin{itemize}
\item 
We propose a simple yet effective learning method that is robust against labeling errors. The proposed SRT method generates the temporal self-ensemble networks by adopting the learning rate scheduling used for SWA.  Without training multiple networks, our approach effectively mitigates the self-bias problem that arises when filtering out noisy labels.
\item We construct a new acquisition criterion for detecting accurately labeled samples using the temporal self-ensemble networks. We transform the input in different ways, and measure the consistency between multi-view predictions generated by these transformed inputs for each self-ensemble network. 
 While the temporal self-ensemble captures the posterior under model uncertainty, the multi-view predictions capture that under data uncertainty. Previous works have not considered both of these uncertainties together to robustly train a model with noisy labels, and this is how our method yields a superior performance than the existing methods. 
\item
We found a relevant work O2U-Net,  \cite{huang2019o2u} that employed the cyclical LR for robust training.  O2U-Net generates the loss trajectory over epochs using the cyclical LR and ranks the samples based on the averaged loss. It filters the noisy samples only after the cyclical training step is complete.  This is contrasted with our SRT, which filters the noisy samples while training the model. While O2U-Net employs the cyclical LR to differentiate between noisy samples and clean samples better,  SRT uses the cyclical LR to produce temporal ensembles.  After cleaning up the noisy samples, O2U-Net needs to go through re-training phase, which requires much longer training time.   Our experimental results confirm that the online filtering approach of  SRT achieves better performance than the post-filtering approach of O2U-Net when self-confirmation issue is well controlled.
\item The source code will be publicly available.
\end{itemize}

\section{Related Work}
To deal with the problem of noisy labels, there have been attempts to correct the loss function such that the impact of noisy labels is minimized.  Several works tried to estimate the noise transition matrix and correct the training loss accordingly \cite{patrini2017fcorrection,natarajan2013learning,xia2019anchor}. 
However, the performance of these methods is sensitive to the quality of the noise transition matrix estimate, in particular, on the dataset with a large number of classes. Other methods proposed a noise-tolerant loss function which penalized the loss terms associated with the misclassified noisy samples \cite{ren2018learning,liu2015classification,wang2017multiclass,reed2014bootstrap}. 

Recent works have actively studied the sample selection strategy which filters the mislabeled data according to the loss value of each sample. MentorNet \cite{jiang2018mentornet} selects the small loss data as correctly labeled data. This is based on the observation that memorization occurs progressively for noisy labels after the DNN learns easy patterns during the initial training phase \cite{arpit2017closer,liu2020earlylearning}. However, self-paced learning \cite{jiang2018mentornet} that selects the small loss data by the network itself causes {\it{self-bias}}. Due to memorization effect, the errors made in sample selection affect the next phase of training without the ability to correct them \cite{bo2018coteaching}. 
To overcome this issue, many studies have used the co-training scheme, which employs two independent networks trained simultaneously to find the noisy labels  \cite{bo2018coteaching,yu2019coteaching+,wei2020jocor,li2020dividemix}. The co-training scheme uses the samples selected by the one network to train the other network to reduce error propagation \cite{bo2018coteaching}. Based on co-training scheme, Co-teaching+ \cite{yu2019coteaching+} adopts the disagreement strategy to make those two networks decoupled. On the other hand, JoCoR \cite{wei2020jocor} adopts the agreement strategy which trains the samples whose predictions from both networks are similar.

The proposed method, SRT treats self-bias issue without co-training scheme. This is possible through using the temporal self-ensemble generated by SGD optimization and utilizing the contrastive loss derived from multi-view predictions to identify incorrectly labeled samples.
Consequently, the proposed SRT outperforms the existing state-of-the-art methods.

\begin{algorithm}
\caption{Summary of SRT algorithm}
\label{algorithm:SRT2}
\begin{algorithmic}[1]
\REQUIRE Training dataset $\mathcal{D}$, the total number of cycles $K$, learning rate cycle $C$ (in epochs), the number of the self-ensemble networks $P'$, the memory $\mathcal{Z}$. \\ 
\STATE Initialize $f(x,\Theta_0)$\\
\FOR{cycle $k = 1, ..., K$}
    \FOR{epoch $t = C*(k-1)+1,..., C*k$}
        \FOR{iter $j = 1,..., \frac{N}{|\mathcal{B}|}$}
            \STATE Update the learning rate  $\epsilon(t, j)$.
            \STATE Sample a mini-batch ${\mathcal{B}}$ from the dataset $\mathcal{D}$
                \STATE Calculate 
                $\mathcal{L}_{filt}(x_i, y_i) $ with $P = \min\{k-1, P'\}$, $\forall (x_i, y_i)\in\mathcal{B}$
            \STATE Filter the noisy labels \\
            ${\mathcal{B}}'$ = $\underset{\tilde{\mathcal{B}}:|\tilde{\mathcal{B}}|\geq{r(t)|{{\mathcal{B}}}}|}{\arg\min}$ $\sum_{(x_i,y_i)\in \tilde{\mathcal{B}}} \mathcal{L}_{filt}(x_i, y_i)$. \\
            \STATE Update the model weights using the training samples in ${\mathcal{B}}'$.
        \ENDFOR
    \ENDFOR
    \STATE Store $f(x; \Theta_k)$ in the memory $\mathcal{Z}$\\
\ENDFOR
\RETURN Trained model  \\
\end{algorithmic}
\end{algorithm}
\section{Proposed Robust Training Method}
In this section, we present the details of the proposed SRT method. 
\subsection{Problem Description}
Consider an $M$-class classification task.
We train the DNN model using a training dataset of size $N$, $\mathcal{D} = \{x_i, y_i\}^{N}_{i=1}$, where $x_i$ is the $i$-th training sample, and $y_i = [y_{i}^{1},...,y_{i}^{M}]^{T}\in\{0, 1\}^{M}$ is the $i$-th label encoded by a one-hot vector.  The DNN model is expressed as $f(x;\Theta)=[f^{1}(x;\Theta), ..., f^{M}(x;\Theta)]^T$, where  $x$ is the input data and $\Theta$ denotes the model parameters. Assuming that some of the samples in $\mathcal{D}$ are corrupted by labeling errors, we aim to filter them in the training phase. Specifically, we intend to design a suitable criterion for identifying correctly labeled samples in the mini-batch, $\mathcal{B} (\subset \mathcal{D})$.

\subsection{Proposed SRT Method }
The structure of the SRT is depicted in Fig. \ref{fig:overall_framwork}.
In this subsection, we will explain the main components of the SRT in detail.
\subsubsection{Generation of temporal self-ensemble}

\input{table/main_results}

\input{table/clothing1M_result}
\input{table/o2u_new}

We generate the temporal self-ensemble while training a single neural network $f(x;\Theta)$ using SGD. We adopt the learning rate scheduling proposed in \cite{loshchilov2016sgdr} and captures the network parameters at the intermediate check-points of SGD iterations. We start with the inital learning rate during a warm-up period of $C$ epochs. Then, we periodically increases the LR rapidly and  decreases it gradually. The LR change according to the number of epochs is illustrated in Fig.~\ref{fig:overall_framwork}.
The LR $\epsilon(t, j)$, specified by the maximum LR $\epsilon_1$, minimum LR $\epsilon_2$, and cycle  $C$ in the number of epochs, is expressed as \cite{loshchilov2016sgdr} 
\begin{align}
    \epsilon(t, j) &= \epsilon_2 + \frac{1}{2}\left(\epsilon_1 - \epsilon_2\right)\left(1+ \cos\left(\frac{s(t, j)}{C}\pi \right)\right),
    \label{eq:cyclical learning rate}
\end{align} 
where 
\begin{align}
   s(t, j) &= \mod(t-1, C) + \frac{|\mathcal{B}|}{N}(j-1),
\end{align}    
for  $t$ and $j$ are the epoch and iteration indices, respectively.   
Thus, whenever $\epsilon(t, j)$ reaches the minimum value $\epsilon_2$, we capture the weights of a temporal self-ensemble network. 
By storing the weights in each LR cycle, the proposed method generates a series of  temporal self-ensemble networks $f(x; \Theta_1), f(x; \Theta_2), ..., f(x; \Theta_{k})$,  where $k$ denotes the LR cycle index.

\subsubsection{Generation of multi-view predictions}
Multi-view predictions are generated by applying data transformation $g(x)$ to the input data $x$ in the mini-batch $\mathcal{B}$.  Two different transformed samples $g(x)$ and $g'(x)$ are generated, where $g(x)$ implies an augmented view of the sample, $x$ and $g'(x)$ implies another view of the same sample, $x$. Note that $g(x)$ and $g'(x)$ are determined by a parameter that defines the  transformation function. We provide the examples of such a transformation function in subsection 4.1. These transformed samples are input to the self-ensemble model $f(x;\Theta_k)$ to produce the multi-view predictions, $f(g(x); \Theta_k)$ and $f(g'(x); \Theta_k)$.

\subsubsection{Filtering the samples with noisy labels}
The proposed SRT method uses the {\it sample acquisition function} to identify a sample with noisy label in the mini-batch. The role of the acquisition function is to generate the likelihood of an input sample being mislabeled. We construct the sample acquisition function using the latest $P$ temporal self-ensemble networks $f(x; \Theta_{k-P}), ..., f(x; \Theta_{k-1})$ obtained up to the previous LR cycles. 
The acquisition function for the given training sample $(x_i, y_i)$ is expressed as
\begin{align}
   & \mathcal{L}_{filt}(x_i,y_i) = \sum_{p=1}^P \lambda \mathcal{L}_{CE}(f(x_i;\Theta_{k-p}), y_i) \nonumber \\ & + (1 - \lambda) \mathcal{L}_{JS}(f(g(x_i);\Theta_{k-p}),f(g'(x_i);\Theta_{k-p})).
    \label{eq:filtering objective}
\end{align}
Note that the individual terms of the summation in (\ref{eq:filtering objective}) are associated with the latest $P$ self-ensemble networks. We do not include the model under training to avoid the self-bias issue. Each term consists of two terms; 1) the cross-entropy (CE) loss term and 2) the Jensen-Shannon (JS) divergence loss term.  $\lambda$ is a parameter that balances the two terms.  The CE loss determines how confident the self-ensemble model is regarding the relationship between  input $x_i$ and label $y_i$
\begin{align}
      \mathcal{L}_{CE}(f(x_i;\Theta_{k-p}), y_i) = - \sum_{m=1}^{M} y_{i}^{m}\log{f^m(x_i;\Theta_{k-p})}.
      \label{eq:cross-entropy objective} 
\end{align}
 The CE loss term is likely to be higher for incorrect labels.
The second term measures the consensus of the multi-view predictions $g(x_i)$ and $g'(x_i)$ through the JS divergence 
\begin{align}
      &\mathcal{L}_{JS}(f(g(x_i);\Theta_{k-p}), f(g'(x_i);\Theta_{k-p})) \nonumber \\&= D_{KL}(f(g(x_i);\Theta_{k-p})|f(g'(x_i); \Theta_{k-p})) \nonumber \\
      & \quad + D_{KL}(f(g'(x_i);\Theta_{k-p})|f(g(x_i);\Theta_{k-p})),
      \label{eq:Jenson-shannon objective} 
\end{align}
where
$ D_{KL}(p|p') = \sum_{m=1}^{M} p^m \log{ \left(\frac{p^m }{p'^m }\right)}$ and $p=[p^1,...,p^M]^{T}$. We use the simplified form of the JS divergence proposed in \cite{wei2020jocor} for reducing the computation. The multi-view predictions are more likely to fluctuate for the mislabeled samples than for the correctly labeled samples \cite{lee2020ltec}. Thus, the JS divergence loss term tends to be large for the mislabeled samples and helps detecting these samples. Fig. \ref{fig:js} shows the values of JS loss averaged over noisy, clean, total samples separately for each epoch on CIFAR-10 dataset. The JS loss value tends to be larger for the noisy samples than for the clean samples and the gap between two increases with epoch.

The proposed method computes the acquisition function for all samples in the mini-batch and uses only $r(t)$ \% of samples with the smallest acquisition value for training, where $t$ is the epoch index. Further, we assume that $\eta$ \% of the dataset is corrupted by labeling errors, where the noise rate $\eta$ can be determined from the known statistics or empirical estimation. At the beginning of training, $r(t)$ is set slightly larger than $(100-\eta)$ \% to avoid filtering the clean samples and then gradually reduces to $(100-\eta)$ \% as the model improves with training epochs \cite{bo2018coteaching}. 
 

\subsubsection{Training with filtered samples}
We consider $k$ as the current LR cycle index and  $\mathcal{B}'$ as the set of filtered samples in the mini-batch. The SRT method trains the model by minimizing the following loss function
\begin{align}
    &\mathcal{L}_{train}(x_i, y_i) = \frac{1}{|\mathcal{B'}|}\sum_{(x_i, y_i) \in \mathcal{B}'} \lambda \mathcal{L}_{CE}(f(x_i;\Theta), y_i) \nonumber \\ & \quad \quad \quad + (1 - \lambda) \mathcal{L}_{JS}(f(g(x_i);\Theta), f(g'(x_i);\Theta)),
    \label{eq:training objective} 
\end{align}
where $f(x;\Theta)$ denotes the model under training. Aside from the filtering step, the CE loss  and JS divergence loss terms are used to train  the model. The JS divergence loss term for the training loss corresponds to the {\it consistency regularization}  used for semi-supervised learning \cite{laine2017temporal}. While the CE loss is used to learn from the target label, the JS divergence loss regularizes the model to make consistent predictions for the perturbed inputs on the same data. The combination of these two loss terms enables an improvement in classification performance. A back-propagation algorithm is executed to minimize the loss function $\mathcal{L}_{train}$.

\subsubsection{Summary of SRT algorithm}
Algorithm \ref{algorithm:SRT2} summarizes the detailed procedure of the proposed SRT method.

\section{Experiments}
In this section, we evaluate the performance of the proposed SRT method. 

\subsection{Experimental Setup}
\subsubsection{Datasets}
We used four popular public datasets, MNIST \cite{lecun1998mnist}, CIFAR-10, CIFAR-100 \cite{krizhevsky2009learning}, and Clothing1M  \cite{xiao2015clothing1m} for the evaluation.
For MNIST, CIFAR-10, and CIFAR-100, both symmetric and asymmetric label noises were synthetically generated. We followed the recommendations from previous studies \cite{bo2018coteaching, wei2020jocor} to generate synthetic label noise. Clothing1M \cite{xiao2015clothing1m} is a large-scale dataset containing real-world noisy labels. The annotation was conducted by crawling online websites.

\subsubsection{Evaluation}

The performance was evaluated by measuring the test accuracy achieved by the models for each dataset. The test accuracy was averaged over the last 10 epochs. The  mean and standard deviation of the test accuracy were presented by repeating the experiments five times with different random seeds.  To evaluate the accuracy of the filtering methods, we also measured the {\it label precision}, which is defined as the average ratio of the clean samples to the selected samples in the mini-batch \cite{bo2018coteaching}.

We compared the proposed SRT method with the existing methods including Standard (vanilla supervised learning with noisy labels), Decoupling \cite{malach2017decoupling},  Co-teaching \cite{bo2018coteaching}, Co-teaching+ \cite{yu2019coteaching+}, JoCoR \cite{wei2020jocor}, and Jo-SRC \cite{yao2021josrc}. We reproduced the performance of these methods except for Jo-SRC. The performance of Jo-SRC was brought from the original paper\footnote{In our experiment, we could not reproduce the performance of Jo-SRC.}. We also compared SRT with O2U-Net, which used cyclic LR for different purpose.  Because the training configurations of O2U-Net were provided for different network architecture (e.g., nine-layer convolutional neural network), we presented this comparison separately in Table \ref{table:o2u}.  Note that we compared SRT with the ensemble-based robust training methods and did not include  the semi-supervised learning-based methods \cite{li2020dividemix, song2019selfie, nguyen2020self, yao2021josrc, jiang2020synthetic, Kim_2021_CVPR, zhou2021robust} for comparison since they used unlabeled data samples for training.

\subsubsection{Experimental setup}

The experimental setup of previous studies  was employed \cite{wei2020jocor}. For CIFAR-10 and CIFAR100, we used a seven-layer convolutional neural network with a batch size of 128. For MNIST, we used a two-layer perceptron with a batch size of 128. For Clothing1M, we used ResNet-18 \cite{he2016deep} with a batch size of 64. The parameter $\lambda$ in (\ref{eq:training objective}) was set to 0.9 for MNIST and was set to 0.7 for CIFAR-10, CIFAR-100, and Clothing1M. For all experiments, we used the Adam optimizer with a momentum of 0.9 and set $P=2$ for sample selection. Except for Clothing1M, we set $\epsilon_1 = 0.001$, $\epsilon_2=0$, and $C=20$ epochs for cyclical LR scheduling in (\ref{eq:cyclical learning rate}). For Clothing1M, we set $\epsilon_1 = 0.0005$, $\epsilon_2=0$, and $C=10$ epochs. Following Co-teaching \cite{bo2018coteaching}, we used $r(t) = 1 - \frac{\eta}{100} \cdot \min \{\frac{t}{T_k}, 1\}$, where $t$ is the epoch index and $T_k$ is the hyper-parameter. We set $T_k=10$ for MNIST, $T_k=15$ for CIFAR-10 and CIFAR-100, and $T_k=5$ for Clothing1M. Each $T_k$ was empirically determined. We conducted all experiments using a TITAN XP GPU.

Data transformations used to generate multi-view predictions include random translation, horizontal flipping, random cropping, and normalization \cite{shorten2019survey}. We applied a different set of data transformations for each dataset. For CIFAR-10 and CIFAR-100, we used a combination of random translation by up to four pixels, horizontal flipping, and normalization \cite{he2016deep}. For MNIST, we used random translation by up to one pixel followed by normalization \cite{visin2015renet}. For Clothing1M, we performed horizontal flipping, resizing to 256$\times$256, and random cropping to 224$\times$224, followed by normalization \cite{wei2020jocor}. 

\begin{figure*}[t]
    \centering
    \includegraphics[width=0.95\textwidth]{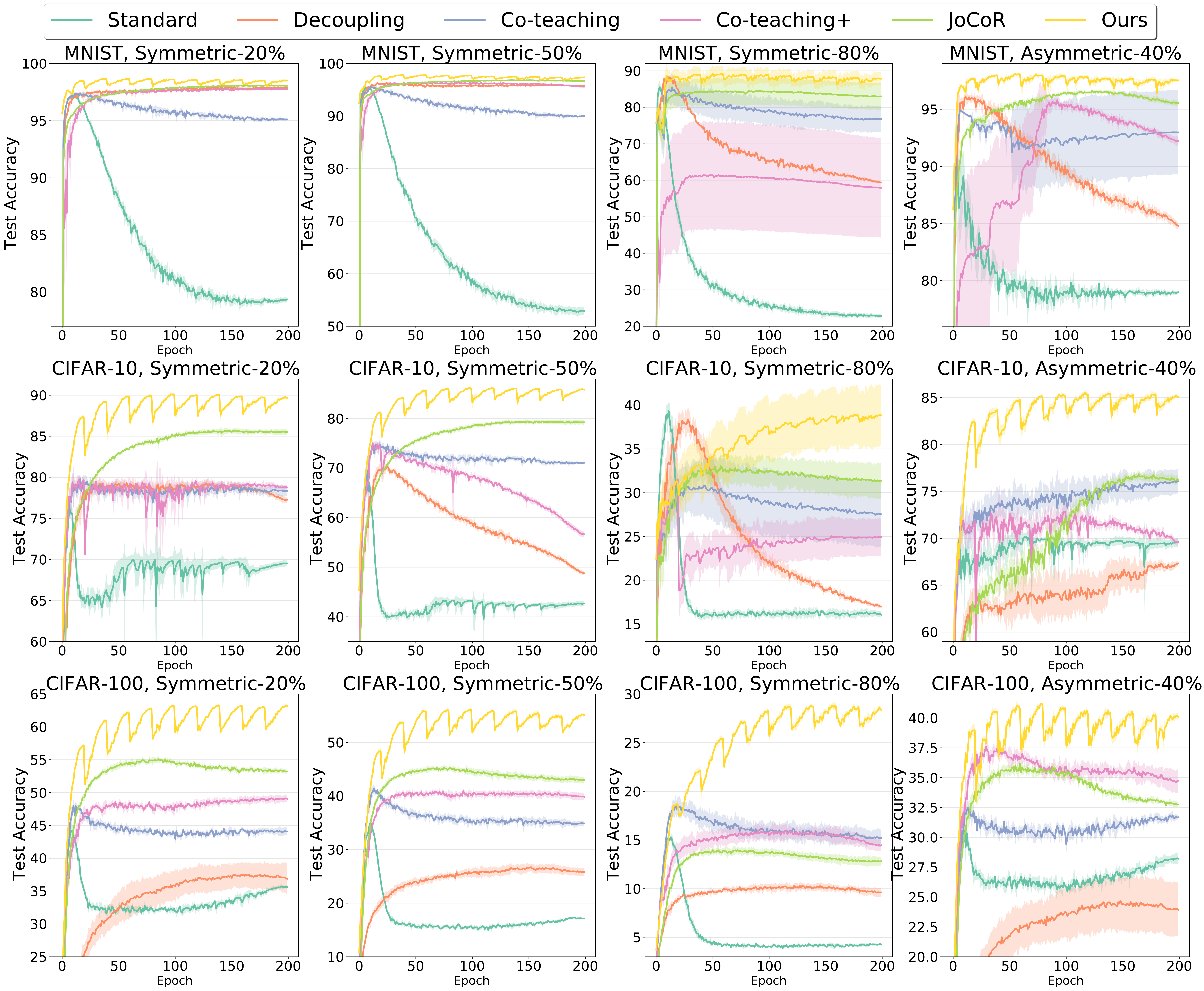}
    \caption{Test accuracy versus epoch index evaluated on MNIST, CIFAR-10 and CIFAR-100 datasets.}
    \label{fig:test_accuracy}
\end{figure*}

\subsection{Performance Comparison}
Table \ref{table:main results} presents the  performance comparison of several robust training methods evaluated on MNIST, CIFAR-10, and CIFAR-100.
We consider four combinations of different noise types and ratios, i.e., symmetric-20\%, symmetric-50\%, symmetric-80\%, and asymmetric-40\%. The proposed SRT consistently achieves a better performance than the existing robust training methods for all setups considered. The performance gain of the SRT increases with increase in the noise ratio. This shows that the SRT is particularly better at coping with severe labeling errors than the existing methods.
On the MNIST dataset, the SRT method outperforms the second best, JoCoR by 0.45\% in the symmetric-20\% category. The performance gain of the SRT goes up to 5.56\%  for more difficult symmetric-80\% setup.  The SRT exhibits higher performance gains on CIFAR-10 and CIFAR-100 datasets.   In the symmetric-80\% category, the SRT achieves up to 7.37\% and 15.31\% performance gains over the JoCoR on CIFAR-10 and CIFAR-100, respectively. Furthermore, SRT beats the current state-of-the-art, Jo-SRC, by up to 4.42\% and 4.35\% on the same noise settings.

Table \ref{table:Clothing1M} presents the performance evaluated on Clothing1M, the real-world dataset. We followed the evaluation protocol presented in \cite{wei2020jocor}. {\it Best} indicates the achieved test accuracy when the best validation accuracy is observed, and {\it last} indicates the test accuracy obtained after the training is complete. SRT achieves the best performance among the candidate methods. In particular, SRT outperforms the second best, Jo-SRC by  0.70\% in {\it best accuracy}.

 In Table \ref{table:o2u}, we also compare the performance of SRT and O2U-Net. We used a nine-layer convolutional neural network to match the configurations used in O2U-Net. We observe that SRT significantly outperforms O2U-Net on both CIFAR-10 and CIFAR-100 datasets. Note that the performance gap increases with noise ratio, which confirms that SRT is particularly strong for harsh noisy label scenarios.

\input{table/cifar10_ablation_50}
\input{table/js_loss_ablation}

\subsection{Performance Analysis}

In this subsection, we investigate the behavior of the SRT with  training epochs.  We measure the intermediate test accuracy of several algorithms after optimization in each epoch. Fig. \ref{fig:test_accuracy} depicts the variation of test accuracy in epoch.
{\it Standard} is the baseline without any noisy label filtering. During the initial training phase, the test accuracy of {\it Standard} rapidly increases; however, it drastically decays at a certain point as the incorrect labels are memorized by the model. Therefore, an effective training method that can alleviate this memorization effect is necessary.
The robust training methods successfully compensate the performance degradation exhibited in the {\it Standard}. Nevertheless,  Co-teaching, Decoupling,  Co-teaching+, and JoCoR exhibit inconsistent performances in that none of them consistently outperforms the rest.  While JoCoR achieves the best performance for the symmetric-20\% and symmetric-50\% settings, it gets worse than other methods for harder settings. In contrast, the SRT method outperforms the other methods for all setups considered. In most cases, the test accuracy of the SRT does not decay over the entire epoch and maintains a high level of test accuracy. The performance gain of the SRT method increases for more difficult noise settings, which is consistent with the results in Table \ref{table:main results}.  The slight fluctuations in the test accuracy observed for SRT appears to be affected by the cyclical LR scheduling.

\input{table/cifar10_selection_sym80}

\subsection{Ablation Study}

In this section, we conduct an ablation study to analyze the contribution of the proposed concepts behind the SRT method.  Table \ref{table:cifar10_ablation} presents the test accuracy obtained by adding each idea to the baseline for Symmetric-50\% on CIFAR-10 dataset.
{\it Temporal Ensemble} indicates temporal self-ensemble, and {\it Multi-view} indicates the multi-view prediction. Baseline was chosen as a method filtering noisy labels with small loss samples. Since {\it Multi-view} requires a data augmentation strategy, we also present the performance of the baseline with data augmentation enabled.  {\it Temporal Ensemble} improves the mean accuracy by 16.54\% over the baseline. {\it Multi-view} yields 16.62\% improvement. With both {\it Temporal Ensemble} and {\it Multi-view} enabled, the overall performance gain goes up to  18.56\%.

Since the JS divergence term is also used for the training, it is worth taking a look at the effect of the JS divergence loss used for training. Table \ref{table:js} shows how much the classification accuracy increases by adding the JS divergence loss to the CE loss.  We see that the JS divergence loss provides a significant improvement in classification accuracy for all settings considered. In particular, the JS divergence loss achieves 1.88\% performance improvement in Symmetric-50\% on CIFAR-10.


Next, we investigate how the performance varies depending on the way the sample acquisition function is constructed based on the temporal self-ensemble. Table \ref{table:cifar10_selection_sym80} presents the test accuracy achieved when  different combinations of the ensemble networks $f(x;\Theta_{k-2})$ and $f(x;\Theta_{k-1}),$ and the current model $f(x;\Theta_{k})$ are used for the acquisition function. For fair comparison, configurations of cyclical LR and multi-view prediction are equally set for all cases. We consider symmetric-80\% noise setting on CIFAR-10 dataset.  The method (a) is the baseline that filters noisy labels based on small loss samples. Note that the method (c) corresponds to SRT in that it uses the latest two self-ensemble networks. Obviously, we observe that the method (a) suffers from the self-bias issue.
As   the methods (b) and (c) use temporal ensemble networks to filter noisy labels, their performances improve over the method (a). We also note that the method (c) using two  self-ensemble networks is better than the method (b) using a single network. Since the method (d) includes the current model $f(x;\Theta_{k})$ in the acquisition function, it might suffer from self-bias issue.  Table \ref{table:cifar10_selection_sym80} shows that the method (d) performs worse than the method (c), SRT in terms of mean accuracy.

\section{Conclusions}

In this paper, we proposed a simple yet effective method for robustly training a DNN model using mis-labeled data samples. The proposed SRT method filtered out mislabeled samples based on new acquisition criteria derived from temporal self-ensembles.  These self-ensemble networks were obtained by periodically sampling the model weights over the SGD weight trajectory during training. The cyclical LR scheduling, which was popularly used for SWA \cite{izmailov2018averaging}, was adopted to generate better self-ensemble networks.  Furthermore, we added the additional acquisition metric that measures the inconsistency between the multi-view predictions. These predictions were obtained by feeding the transformed inputs into each self-ensemble. Combining the above two components, the SRT filtered out the mislabeled samples and trained the model using the filtered data samples.
The experiments conducted on the widely used public datasets demonstrated that the proposed SRT method offered significant performance gains over the existing methods. 

\bibliography{aaai22}

\end{document}

%% file: table/main_results.tex
\begin{table*}[t]
\centering
  
  \begin{adjustbox}{center}
  \begin{tabular}{c|c|cccc}
    
    \hline
    \hline
    Dataset & Method & Symmetric-20\% & Symmetric-50\% & Symmetric-80\% & Asymmetric-40\%\\
    \hline
    \hline
    \multirow{6}{*}{MNIST}& Standard & 79.29 $\pm$ 0.23 & 52.86 $\pm$ 0.78 & 22.89 $\pm$ 0.50 & 78.94 $\pm$ 0.17 \\
    & Decoupling & 97.85 $\pm$ 0.11 & 95.74 $\pm$ 0.18 & 59.64 $\pm$ 0.36 & 85.13 $\pm$ 0.35 \\
    & Co-teaching & 95.12 $\pm$ 0.14 & 89.97 $\pm$ 0.17 & 76.78 $\pm$ 3.60 & 92.97 $\pm$ 3.68 \\
    & Co-teaching+ & 97.75 $\pm$ 0.08 & 95.65 $\pm$ 0.21 & 58.03 $\pm$ 13.68 & 92.34 $\pm$ 0.35 \\
    & JoCoR & 98.06 $\pm$ 0.06 & 96.68 $\pm$ 0.10 & 83.01 $\pm$ 3.64 & 95.57 $\pm$ 0.23 \\
    & \bf{SRT(ours)} & \bf{98.51} $\pm$ 0.11 & \bf{97.37} $\pm$ 0.06 & \bf{88.57} $\pm$ 0.68 & \bf{97.51} $\pm$ 0.11 \\
    \hline
    \multirow{7}{*}{CIFAR-10}&Standard & 69.46 $\pm$ 0.36 & 42.62 $\pm$ 0.51 & 16.18 $\pm$ 0.36 & 69.50 $\pm$ 0.44 \\
    & Decoupling & 77.39 $\pm$ 0.41 & 49.22 $\pm$ 0.44 & 17.13 $\pm$ 0.21 & 67.11 $\pm$ 0.37 \\
    & Co-teaching & 78.36 $\pm$ 0.62 & 71.04 $\pm$ 0.24 & 27.59 $\pm$ 3.76 & 76.00 $\pm$ 1.32 \\
    & Co-teaching+ & 78.81 $\pm$ 0.26 & 57.12 $\pm$ 0.59 & 24.90 $\pm$ 2.13 & 69.88 $\pm$ 0.33 \\
    & JoCoR & 85.52 $\pm$ 0.34 & 79.24 $\pm$ 0.47 & 31.35 $\pm$ 2.01 & 76.26 $\pm$ 0.49 \\
    & Jo-SRC & 87.80 $\pm$ 0.30 & 73.67 $\pm$ 0.09 & 34.30 $\pm$ 2.30 & 79.96 $\pm$ 0.17 \\
    & \bf{SRT(ours)} & \bf{89.53} $\pm$ 0.13 & \bf{85.66} $\pm$ 0.08 & \bf{38.72} $\pm$ 3.52 & \bf{84.96} $\pm$ 0.21 \\
    \hline
    \multirow{7}{*}{CIFAR-100} &Standard & 35.43 $\pm$ 0.42 & 17.30 $\pm$ 0.28 & 4.26 $\pm$ 0.09 & 28.06 $\pm$ 0.42 \\
    & Decoupling & 36.54 $\pm$ 2.34 & 25.43 $\pm$ 1.11 & 9.81 $\pm$ 0.50 & 24.27 $\pm$ 2.12 \\
    & Co-teaching & 44.08 $\pm$ 0.54 & 34.88 $\pm$ 0.47 & 15.12 $\pm$ 0.78 & 31.75 $\pm$ 0.28 \\
    & Co-teaching+ & 49.01 $\pm$ 0.49 & 39.74 $\pm$ 0.74 & 14.40 $\pm$ 0.58 & 34.75 $\pm$ 0.86 \\
    & JoCoR & 53.21 $\pm$ 0.34 & 42.84 $\pm$ 0.64 & 12.84 $\pm$ 0.43 & 32.80 $\pm$ 0.21 \\
    & Jo-SRC & 58.15 $\pm$ 0.14 & 51.26 $\pm$ 0.11 &  23.80 $\pm$ 0.05 & 38.52 $\pm$ 0.20\\
    & \bf{SRT(ours)} & \bf{62.77} $\pm$ 0.15 & \bf{54.81} $\pm$ 0.35 & \bf{28.15} $\pm$ 0.31 & \bf{40.04} $\pm$ 0.20 \\
    \hline
    \hline
  \end{tabular}
  \end{adjustbox}
  \caption{Average test accuracy (\%) on MNIST, CIFAR-10 and CIFAR-100 datasets.} 
  \label{table:main results}
\end{table*}

%% file: table/clothing1M_result.tex
\begin{table}[tbh]
    \centering
    \begin{adjustbox}{center}
    \begin{tabular}{c|cc}
    \hline
    \hline
    Methods & {\it best} & {\it last} \\
    \hline
    \hline
    Standard & 67.22 & 64.68 \\
    Decoupling & 68.48 & 67.32 \\
    Co-teaching & 69.21 & 68.51\\
    Co-teaching+ & 59.32 & 58.79 \\
    JoCoR & 70.30 & 69.79 \\
    Jo-SRC & 71.78 & -\\
    \hline 
    SRT & \bf{72.48} $\pm$ 0.26 & \bf{72.47} $\pm$ 0.21 \\
    \hline
    \hline
    \end{tabular}
    \end{adjustbox}
    \caption{Average test accuracy (\%) on Clothing1M.}
    \label{table:Clothing1M}
\end{table}

%% file: table/o2u_new.tex
\begin{table*}[t]
\centering

  \begin{adjustbox}{center}
  \begin{tabular}{c|c|ccc}

    \hline
    \hline
    Method & Dataset & Symmetric-10\% & Symmetric-20\% & Symmetric-40\%\\

    \hline
    \multirow{2}{*}{O2U-Net} & CIFAR-10 & 87.64\% & 85.24\% & 79.64\% \\
     & CIFAR-100 & 62.32\% & 60.53\% & 52.47\% \\
    \hline
    \multirow{2}{*}{SRT} & CIFAR-10 & \bf{92.11\%} & \bf{91.64\%} & \bf{88.94\%} \\
    
    & CIFAR-100 & \bf{66.33\%} & \bf{64.17\%} & \bf{57.78\%} \\

    \hline
    \hline
  \end{tabular}
  \end{adjustbox}
  \caption{Comparison with O2U-Net on CIFAR-10 and CIFAR-100.}
  \label{table:o2u}
\end{table*}

%% file: table/cifar10_ablation_50.tex

  
  

\begin{table*}[tbh]

\centering
  
  \begin{adjustbox}{center}
  \begin{tabular}{c|c|c|c}
  \hline
  \hline 
  \multirow{2}{*}{Method} & \multicolumn{2}{c|}{Components} &  Noise Rate\\
  \cline{2-4} & Temporal Ensemble & Multi-view &  Symmetric-50\% \\
  \hline
  Baseline & & & 63.80 $\pm$ 0.07 \\
    Baseline + Data augmentation & & & 67.10 $\pm$ 0.37 \\
  \hline
  \multirow{3}{*}{SRT}& $\checkmark$ & & 83.64 $\pm$ 0.17 \\
  & & $\checkmark$ & 83.72 $\pm$ 0.05 \\
  & $\checkmark$ & $\checkmark$ & \bf{85.66} $\pm$ 0.08 \\
  \hline
  \hline
  \end{tabular}
  \end{adjustbox}
  \caption{Ablation study evaluated on CIFAR-10. }
  \label{table:cifar10_ablation}
\end{table*}

%% file: table/js_loss_ablation.tex
\begin{table*}[tbh]

\centering
  
  \begin{adjustbox}{center}
  \begin{tabular}{c|c|cccc}
  \hline
  \hline 
  Dataset & Method & Symmetric-20\% & Symmetric-50\% & Symmetric-80\% & Asymmetric-40\% \\
  \hline
  \multirow{2}{*}{CIFAR-10} & CE & 88.26 $\pm$ 0.21 & 83.78 $\pm$ 0.27 & 36.36 $\pm$ 4.48 & 83.41 $\pm$ 0.22 \\
  & CE  + JS & \bf{89.53} $\pm$ \bf{0.13}  & \bf{85.66} $\pm$ \bf{0.13}  & \bf{38.72} $\pm$ \bf{3.52}  & \bf{84.96} $\pm$ \bf{0.21}  \\
  \hline
  \multirow{2}{*}{CIFAR-100} & CE & 60.32 $\pm$ 0.16 & 53.41 $\pm$ 0.75 & 27.44 $\pm$ 1.44 & 37.14 $\pm$ 0.21 \\
  & CE + JS & \bf{62.77} $\pm$ \bf{0.15}  & \bf{54.81} $\pm$ \bf{0.35}  & \bf{28.15} $\pm$ \bf{0.31}  & \bf{40.04} $\pm$ \bf{0.20}  \\

  \hline
  \hline
  \end{tabular}
  \end{adjustbox}
  \caption{Effect of JS divergence loss term for training.}
  \label{table:js}
\end{table*}

%% file: table/cifar10_selection_sym80.tex
\begin{table*}[t]
\centering
  
  \begin{adjustbox}{center}
  \begin{tabular}{c|ccc|c}
    
    \hline
    \hline
    \multirow{2}{*}{Method} & \multicolumn{3}{c|}{Model} & \multirow{2}{*}{Accuracy (\%)}\\
    \cline{2-4}& $f(x;\Theta_{k-2})$ & $f(x;\Theta_{k-1})$ & $f(x;\Theta_{k})$  &\\
    \hline
    (a) && &  $\checkmark$& 36.28 $\pm$ 4.65 \\
    (b) & &$\checkmark$ & & 36.71 $\pm$ 5.24 \\
    (c) &$\checkmark$& $\checkmark$& &\bf{38.72} $\pm$ 3.52\\ 
    (d) &$\checkmark$& $\checkmark$ & $\checkmark$ & 37.45 $\pm$ 3.01\\
    \hline
    \hline
  \end{tabular}
  \end{adjustbox}
  \caption{Comparison of the different methods to construct the acquisition function on CIFAR-10.}
  \label{table:cifar10_selection_sym80}
\end{table*}

%% file: aaai22.bbl
\begin{thebibliography}{35}
\providecommand{\natexlab}[1]{#1}

\bibitem[{Arpit et~al.(2017)Arpit, Jastrz{\k{e}}bski, Ballas, Krueger, Bengio,
  Kanwal, Maharaj, Fischer, Courville, Bengio, and
  Lacoste-Julien}]{arpit2017closer}
Arpit, D.; Jastrz{\k{e}}bski, S.; Ballas, N.; Krueger, D.; Bengio, E.; Kanwal,
  M.~S.; Maharaj, T.; Fischer, A.; Courville, A.; Bengio, Y.; and
  Lacoste-Julien, S. 2017.
\newblock {A Closer Look at Memorization in Deep Networks}.
\newblock In \emph{Proceedings of International Conference on Machine Learning
  (ICML)}, 233--242.

\bibitem[{Han et~al.(2018)Han, Yao, Yu, Niu, Xu, Hu, Tsang, and
  Sugiyama}]{bo2018coteaching}
Han, B.; Yao, Q.; Yu, X.; Niu, G.; Xu, M.; Hu, W.; Tsang, I.~W.; and Sugiyama,
  M. 2018.
\newblock {Co-teaching: Robust Training of Deep Neural Networks with Extremely
  Noisy Labels}.
\newblock In \emph{Advances in Neural Information Processing Systems
  (NeurIPS)}, 8536--8546.

\bibitem[{He et~al.(2016)He, Zhang, Ren, and Sun}]{he2016deep}
He, K.; Zhang, X.; Ren, S.; and Sun, J. 2016.
\newblock {Deep Residual Learning for Image Recognition}.
\newblock In \emph{Proceedings of the IEEE Conference on Computer Vision and
  Pattern Recognition (CVPR)}, 770--778.

\bibitem[{Huang et~al.(2019)Huang, Qu, Jia, and Zhao}]{huang2019o2u}
Huang, J.; Qu, L.; Jia, R.; and Zhao, B. 2019.
\newblock O2u-net: A simple noisy label detection approach for deep neural
  networks.
\newblock In \emph{Proceedings of the IEEE International Conference on Computer
  Vision (ICCV)}, 3326--3334.

\bibitem[{Izmailov et~al.(2018)Izmailov, Podoprikhin, Garipov, Vetrov, and
  Wilson}]{izmailov2018averaging}
Izmailov, P.; Podoprikhin, D.; Garipov, T.; Vetrov, D.; and Wilson, A.~G. 2018.
\newblock Averaging weights leads to wider optima and better generalization.
\newblock \emph{arXiv preprint arXiv:1803.05407}.

\bibitem[{Jiang et~al.(2020)Jiang, Huang, Liu, and Yang}]{jiang2020synthetic}
Jiang, L.; Huang, D.; Liu, M.; and Yang, W. 2020.
\newblock Beyond Synthetic Noise: Deep Learning on Controlled Noisy Labels.
\newblock In \emph{Proceedings of International Conference on Machine Learning
  (ICML)}, 4804--4815.

\bibitem[{Jiang et~al.(2018)Jiang, Zhou, Leung, Li, and
  Fei{-}Fei}]{jiang2018mentornet}
Jiang, L.; Zhou, Z.; Leung, T.; Li, L.; and Fei{-}Fei, L. 2018.
\newblock MentorNet: Learning Data-Driven Curriculum for Very Deep Neural
  Networks on Corrupted Labels.
\newblock In \emph{Proceedings of International Conference on Machine Learning
  (ICML)}, 2309--2318.

\bibitem[{Kim et~al.(2019)Kim, Yim, Yun, and Kim}]{kim2019nlnl}
Kim, Y.; Yim, J.; Yun, J.; and Kim, J. 2019.
\newblock {NLNL:} Negative Learning for Noisy Labels.
\newblock In \emph{Proceedings of the IEEE International Conference on Computer
  Vision (ICCV)}, 101--110.

\bibitem[{Kim et~al.(2021)Kim, Yun, Shon, and Kim}]{Kim_2021_CVPR}
Kim, Y.; Yun, J.; Shon, H.; and Kim, J. 2021.
\newblock Joint Negative and Positive Learning for Noisy Labels.
\newblock In \emph{Proceedings of the IEEE Conference on Computer Vision and
  Pattern Recognition (CVPR)}, 9442--9451.

\bibitem[{Krizhevsky(2009)}]{krizhevsky2009learning}
Krizhevsky, A. 2009.
\newblock Learning Multiple Layers of Features from Tiny Images.
\newblock \emph{Technical report, Department of Computer Science, University of
  Toronto}.

\bibitem[{LeCun et~al.(1998)LeCun, Bottou, Bengio, and
  Haffner}]{lecun1998mnist}
LeCun, Y.; Bottou, L.; Bengio, Y.; and Haffner, P. 1998.
\newblock Gradient-Based Learning Applied to Document Recognition.
\newblock \emph{Proceedings of the IEEE}, 86(11): 2278--2324.

\bibitem[{Lee and Chung(2020)}]{lee2020ltec}
Lee, J.; and Chung, S. 2020.
\newblock Robust training with ensemble consensus.
\newblock In \emph{Proceedings of the International Conference on Learning
  Representations (ICLR)}.

\bibitem[{Li, Socher, and Hoi(2020)}]{li2020dividemix}
Li, J.; Socher, R.; and Hoi, S. C.~H. 2020.
\newblock DivideMix: Learning with Noisy Labels as Semi-supervised Learning.
\newblock In \emph{Proceedings of the International Conference on Learning
  Representations (ICLR)}.

\bibitem[{Liu et~al.(2020)Liu, Niles-Weed, Razavian, and
  Fernandez-Granda}]{liu2020earlylearning}
Liu, S.; Niles-Weed, J.; Razavian, N.; and Fernandez-Granda, C. 2020.
\newblock Early-Learning Regularization Prevents Memorization of Noisy Labels.
\newblock In \emph{Advances in Neural Information Processing Systems
  (NeurIPS)}.

\bibitem[{Liu and Tao(2015)}]{liu2015classification}
Liu, T.; and Tao, D. 2015.
\newblock Classification with Noisy Labels by Importance Reweighting.
\newblock \emph{IEEE Transactions on Pattern Analysis and Machine Intelligence
  (TPAMI)}, 38(3): 447--461.

\bibitem[{Loshchilov and Hutter(2016)}]{loshchilov2016sgdr}
Loshchilov, I.; and Hutter, F. 2016.
\newblock {SGDR:} Stochastic Gradient Descent with Warm Restarts.
\newblock \emph{Learning}, 10: 3.

\bibitem[{Malach and Shalev{-}Shwartz(2017)}]{malach2017decoupling}
Malach, E.; and Shalev{-}Shwartz, S. 2017.
\newblock Decoupling "when to update" from "how to update".
\newblock In \emph{Advances in Neural Information Processing Systems
  (NeurIPS)}, 960--970.

\bibitem[{Natarajan et~al.(2013)Natarajan, Dhillon, Ravikumar, and
  Tewari}]{natarajan2013learning}
Natarajan, N.; Dhillon, I.~S.; Ravikumar, P.~K.; and Tewari, A. 2013.
\newblock Learning with Noisy Labels.
\newblock In \emph{Advances in Neural Information Processing Systems
  (NeurIPS)}, 1196--1204.

\bibitem[{Nguyen et~al.(2020)Nguyen, Mummadi, Ngo, Nguyen, Beggel, and
  Brox}]{nguyen2020self}
Nguyen, D.~T.; Mummadi, C.~K.; Ngo, T.; Nguyen, T. H.~P.; Beggel, L.; and Brox,
  T. 2020.
\newblock {SELF:} Learning to Filter Noisy Labels with Self-Ensembling.
\newblock In \emph{Proceedings of the International Conference on Learning
  Representations (ICLR)}.

\bibitem[{Patrini et~al.(2017)Patrini, Rozza, Menon, Nock, and
  Qu}]{patrini2017fcorrection}
Patrini, G.; Rozza, A.; Menon, A.~K.; Nock, R.; and Qu, L. 2017.
\newblock Making Deep Neural Networks Robust to Label Noise: {A} Loss
  Correction Approach.
\newblock In \emph{Proceedings of the IEEE Conference on Computer Vision and
  Pattern Recognition (CVPR)}, 2233--2241.

\bibitem[{Reed et~al.(2015)Reed, Lee, Anguelov, Szegedy, Erhan, and
  Rabinovich}]{reed2014bootstrap}
Reed, S.; Lee, H.; Anguelov, D.; Szegedy, C.; Erhan, D.; and Rabinovich, A.
  2015.
\newblock Training Deep Neural Networks on Noisy Labels with Bootstrapping.
\newblock In \emph{Proceedings of the International Conference on Learning
  Representations (ICLR)}.

\bibitem[{Ren et~al.(2018)Ren, Zeng, Yang, and Urtasun}]{ren2018learning}
Ren, M.; Zeng, W.; Yang, B.; and Urtasun, R. 2018.
\newblock Learning to Reweight Examples for Robust Deep Learning.
\newblock In \emph{Proceedings of International Conference on Machine Learning
  (ICML)}, 4334--4343.

\bibitem[{Samuli and Timo(2017)}]{laine2017temporal}
Samuli, L.; and Timo, A. 2017.
\newblock Temporal Ensembling for Semi-Supervised Learning.
\newblock In \emph{Proceedings of the International Conference on Learning
  Representations (ICLR)}.

\bibitem[{Shen and Sanghavi(2019)}]{shen2019itlm}
Shen, Y.; and Sanghavi, S. 2019.
\newblock Learning with Bad Training Data via Iterative Trimmed Loss
  Minimization.
\newblock In \emph{Proceedings of International Conference on Machine Learning
  (ICML)}, 5739--5748.

\bibitem[{Shorten and Khoshgoftaar(2019)}]{shorten2019survey}
Shorten, C.; and Khoshgoftaar, T.~M. 2019.
\newblock A Survey on Image Data Augmentation for Deep Learning.
\newblock \emph{Journal of Big Data}, 6(1): 1--48.

\bibitem[{Song, Kim, and Lee(2019)}]{song2019selfie}
Song, H.; Kim, M.; and Lee, J. 2019.
\newblock {SELFIE:} Refurbishing Unclean Samples for Robust Deep Learning.
\newblock In \emph{Proceedings of International Conference on Machine Learning
  (ICML)}, 5907--5915.

\bibitem[{Visin et~al.(2015)Visin, Kastner, Cho, Matteucci, Courville, and
  Bengio}]{visin2015renet}
Visin, F.; Kastner, K.; Cho, K.; Matteucci, M.; Courville, A.; and Bengio, Y.
  2015.
\newblock ReNet: A Recurrent Neural Network Based Alternative to Convolutional
  Networks.
\newblock \emph{arXiv preprint arXiv:1505.00393}.

\bibitem[{Wang, Liu, and Tao(2017)}]{wang2017multiclass}
Wang, R.; Liu, T.; and Tao, D. 2017.
\newblock Multiclass Learning with Partially Corrupted Labels.
\newblock \emph{IEEE Transactions on Neural Networks and Learning Systems
  (TNNLS)}, 29(6): 2568--2580.

\bibitem[{Wei et~al.(2020)Wei, Feng, Chen, and An}]{wei2020jocor}
Wei, H.; Feng, L.; Chen, X.; and An, B. 2020.
\newblock Combating Noisy Labels by Agreement: {A} Joint Training Method with
  Co-Regularization.
\newblock In \emph{Proceedings of the IEEE Conference on Computer Vision and
  Pattern Recognition (CVPR)}, 13723--13732.

\bibitem[{Xia et~al.(2019)Xia, Liu, Wang, Han, Gong, Niu, and
  Sugiyama}]{xia2019anchor}
Xia, X.; Liu, T.; Wang, N.; Han, B.; Gong, C.; Niu, G.; and Sugiyama, M. 2019.
\newblock Are Anchor Points Really Indispensable in Label-Noise Learning?
\newblock In \emph{Advances in Neural Information Processing Systems
  (NeurIPS)}, 6838--6849.

\bibitem[{Xiao et~al.(2015)Xiao, Xia, Yang, Huang, and
  Wang}]{xiao2015clothing1m}
Xiao, T.; Xia, T.; Yang, Y.; Huang, C.; and Wang, X. 2015.
\newblock Learning from Massive Noisy Labeled Data for Image Classification.
\newblock In \emph{Proceedings of the IEEE Conference on Computer Vision and
  Pattern Recognition (CVPR)}, 2691--2699.

\bibitem[{Yao et~al.(2021)Yao, Sun, Zhang, Shen, Wu, Zhang, and
  Tang}]{yao2021josrc}
Yao, Y.; Sun, Z.; Zhang, C.; Shen, F.; Wu, Q.; Zhang, J.; and Tang, Z. 2021.
\newblock Jo-SRC: A Contrastive Approach for Combating Noisy Labels.
\newblock In \emph{Proceedings of the IEEE Conference on Computer Vision and
  Pattern Recognition (CVPR)}, 5192--5201.

\bibitem[{Yu et~al.(2019)Yu, Han, Yao, Niu, Tsang, and
  Sugiyama}]{yu2019coteaching+}
Yu, X.; Han, B.; Yao, J.; Niu, G.; Tsang, I.~W.; and Sugiyama, M. 2019.
\newblock How does Disagreement Help Generalization against Label Corruption?
\newblock In \emph{Proceedings of International Conference on Machine Learning
  (ICML)}, 7164--7173.

\bibitem[{Zhang et~al.(2016)Zhang, Bengio, Hardt, Recht, and
  Vinyals}]{zhang2016understanding}
Zhang, C.; Bengio, S.; Hardt, M.; Recht, B.; and Vinyals, O. 2016.
\newblock Understanding Deep Learning Requires Rethinking Generalization.
\newblock In \emph{Proceedings of the International Conference on Learning
  Representations (ICLR)}.

\bibitem[{Zhou, Wang, and Bilmes(2021)}]{zhou2021robust}
Zhou, T.; Wang, S.; and Bilmes, J. 2021.
\newblock Robust Curriculum Learning: from clean label detection to noisy label
  self-correction.
\newblock In \emph{Proceedings of the International Conference on Learning
  Representations (ICLR)}.

\end{thebibliography}
